\documentclass[conference]{IEEEtran}
\IEEEoverridecommandlockouts
\usepackage{cite}
\usepackage{amsmath,amssymb,amsfonts}
\usepackage{algorithmic}
\usepackage{graphicx}
\usepackage{textcomp}
\usepackage{booktabs}
\usepackage[pdf]{graphviz}
\usepackage{morewrites}
\usepackage{float}
\usepackage{multirow}
\usepackage{xcolor,colortbl}
\usepackage{pgfgantt}
\usepackage{adjustbox}
\usepackage{lscape}
\usepackage{caption}
\usepackage{subcaption}
\usepackage{url}
\def\BibTeX{{\rm B\kern-.05em{\sc i\kern-.025em b}\kern-.08em
		T\kern-.1667em\lower.7ex\hbox{E}\kern-.125emX}}

\graphicspath{{figs/}}

\begin{document}
	
	\title{Optimizing Neural Architecture Search using Limited GPU Time in a Dynamic Search Space: A Gene Expression Programming Approach\
		\thanks{This work was supported by CAPES/CNPq.}
	}
	
	\author{\IEEEauthorblockN{1\textsuperscript{st} Jeovane Honorio Alves}
		\IEEEauthorblockA{\textit{Laboratory of Vision, Robotics and Imaging (VRI)} \\
			\textit{Federal University of Parana (UFPR)}\\
			Curitiba -- PR, Brazil \\
			jhalves@inf.ufpr.br}
		\and
		\IEEEauthorblockN{2\textsuperscript{nd} Lucas Ferrari de Oliveira}
		\IEEEauthorblockA{\textit{Laboratory of Vision, Robotics and Imaging (VRI)} \\
			\textit{Federal University of Parana (UFPR)}\\
			Curitiba -- PR, Brazil \\
			lferrari@inf.ufpr.br}
	}
	
	\maketitle
	
	\begin{abstract}
		Efficient identification of people and objects, segmentation of regions of interest and extraction of relevant data in images, texts, audios and videos are evolving considerably in these past years, which deep learning methods, combined with recent improvements in computational resources, contributed greatly for this achievement. Although its outstanding potential, development of efficient architectures and modules requires expert knowledge and amount of resource time available. In this paper, we propose an evolutionary-based neural architecture search approach for efficient discovery of convolutional models in a dynamic search space, within only 24 GPU hours. With its efficient search environment and phenotype representation, Gene Expression Programming is adapted for network's cell generation. Despite having limited GPU resource time and broad search space, our proposal achieved similar state-of-the-art to manually-designed convolutional networks and also NAS-generated ones, even beating similar constrained evolutionary-based NAS works. The best cells in different runs achieved stable results, with a mean error of 2.82\% in CIFAR-10 dataset (which the best model achieved an error of 2.67\%) and 18.83\% for CIFAR-100 (best model with 18.16\%). For ImageNet in the mobile setting, our best model achieved top-1 and top-5 errors of 29.51\% and 10.37\%, respectively. Although evolutionary-based NAS works were reported to require a considerable amount of GPU time for architecture search, our approach obtained promising results in little time, encouraging further experiments in evolutionary-based NAS, for search and network representation improvements.
	\end{abstract}    
	
	
	\section{Introduction}
	
	In conventional machine learning, feature engineering is employed with an intent to develop feature descriptors for a specific problem -- feature which would represent not only a set of samples, but its overall population.
	Although this proposal is interesting, it is difficult and exhaustive to develop feature descriptors to solve a specific problem.
	Thus, deep learning would come to remove the need for feature engineering -- the weights of a network would be optimized to a dataset for accurate representation.
	Following the idea of automatic feature engineering, automation of network design would be a next step in deep learning -- the arduous job of optimizing the structure of a deep network would fall to a computer method, reducing human workload.
	Not only that, but it may provide efficient networks not followed by human thinking (structures with unconventional patterns).
	This would drive research to a new area: Neural Architecture Search (NAS), a sub-field of AutoML (Automated Machine Learning).
	The main idea of NAS is to obtain optimum deep learning models (convnets, recurrent nets and so on) using another method, rather than a human manually designing the model.
	Thus, this method would search for an architecture that better suits a specific problem.
	This outstanding idea would be applied for a variety of deep networks to solve computer vision, machine translation, speech recognition problems and so on \cite{elsken2018neural}.
	
	In this paper, we employ a methodology based on GEP, entitled as Neural Architecture Search using Gene Expression Programming (NASGEP)\footnote{Code for architecture search and training available at \url{http://github.com/jeohalves/nasgep}.}, to assess the likelihood of fast and efficient architecture search with this evolutionary approach.    
	More specifically, we address specific points which can be considered as features for why use this NAS approach:

	\begin{itemize}
		\item Dynamic search space using the GEP representation, which will increase its space considerably but may provide innovative cells if compared with other search spaces (e.g. NAS, DARTS, etc);
		\item Reusable modules -- being treated as GEP genes, tracking which combination of convolutional blocks improves individuals' fitness;
		\item Weights from the fittest modules, not only from convolutional blocks, are passed on to further generations for cutting search time;
		\item Evolutionary search strategy within 24 GPU hours, aiming to obtain efficient models without excessive resource consumption.
	\end{itemize}

	\section{Related Work}
	
	Development of architecture search approaches can be categorized in three dimensions:  
	
	\begin{itemize}
		\item How to represent candidate networks (search space); \item How to find out better candidates (search strategy);
		\item How to evaluate each candidate (parameter estimation strategy).
	\end{itemize}
	
	To develop a NAS approach, these dimensions need to be fully determined.
	
	\subsection{Search Space}
	
	For the first dimension -- search space, the possible structure of each candidate network needs to be defined.
	Two principal points need to be carefully address: a limited search space size, since extensive number of candidates would hinder optimum size in small search time; but also networks would be different between each other as much as possible, since structure variance would introduce the possibility to find out which structure patterns produce fitter results \cite{elsken2018neural}.
	
	Currently, there are three categories of search spaces: chain-structure, multi-branch  and cell-based ones.
	Chain-based NAS works focus on developing optimum sequential models, although with possible presence of skip-connections.
	Generally, these models are simpler and do not present significant differences between manually-designed models like ResNet, WideResNet and so on.
	
	Unlike them, multi-branch NAS approaches (like \cite{elsken2017simple, elsken1804efficient}) focus on complex optimizations on the entire architecture.
	It is like a big Inception \cite{szegedy2017inception} block as the entire network.
	Being extremely different from chain-based models, this type of model introduces too much complexity, increasing exponentially the search space to be traveled.
	
	As for cell-based ones, they produce superior results in lesser time than the others, as it limits the search space to a part of the network (cells are repeated along the entire network) without reducing considerably the variance between models (a cell has similar structure than multi-branched, although not in size).
	Cell-based search spaces split between normal and reduction cells: the former aims feature representation, being replicated many times through the network; then the latter aims reduction of feature map size (with poolings or convolutions with stride $=2$).
	Thus, the best model always has two different cells (in some cases, the reduction cell is already fixed or it has a similar structure than the normal cell).
	
	Two common cell-based subgroups are used: the NAS-based \cite{elsken2017simple} and the Differentiable Architecture Search (DARTS)-based cells \cite{liu2018darts}, which is basically a directed acyclic graph (DAG).
	Cells based on the NAS work (also NASNet \cite{zoph2017learning} and AmoebaNet \cite{real2018regularized}) are composed of $N$ blocks (generally, $N=4$), in which each block is composed of an addition of two operations (e.g. convolution and/or poolings).
	Each block's output can be used as an input of another block or concatenated with other blocks to generate the cell's output.
	
	In the case of DARTS-based cells, operations are combined based on edges and their combinations (like a directed acyclic graph).
	Assuming there are $M$ sequential edges ($E_0, ..., E_{M-1}$), each edge is combined with previous edges by some operation.
	Then, the edge $E_{M-1}$ combines with all previous edges ($E_0, ..., E_{M-2}$) to produce the cell's output.
	Architecture search for the former aims to select which inputs and operations each block will have; and for the latter, which operations will have between edges.
	Several works based on these two cell representations, such as \cite{elsken2017simple, zoph2017learning, real2018regularized, liu2018darts}, surpassed even human-designed models in CIFAR and ImageNet datasets. 
	
	Although this encourages further usage of them, it also presents a problem regarding human bias.
	These cells present a fixed structure regarding how operations are combined: in the first, always two operations (even if one operation is an identity) are added to generate a block and then these blocks are concatenated to generate the cell's output; in the second, outputs of prior edges are combined in the same way to generate the current edge.
	It reduces drastically what types of structures can be found, also limiting to structures close to models already experimented (introducing human bias).
	Thus, for searching for more different patterns, it is crucial to relax how the cells are generated, however not to a point in which search space becomes too big (e.g. multi-branch).
	
	\subsection{Search Strategy and Parameter Estimation Strategy}
	
	Search strategies in NAS works are mainly in the areas of reinforcement learning (RL), evolutionary computation \cite{ baker2016designing, liu2017hierarchical, real2017large, miikkulainen2017evolving, suganuma2017genetic, zhong2018practical, elsken1804efficient} and gradient-based techniques.
	RL-based NAS would employ different methods such as REINFORCE policy, proximal policy optimization (PPO) and Q-learning (like \cite{zoph2016neural, zoph2017learning, baker2016designing}). The first NAS approaches that achieved state-of-the-art results were based on this type. 
	
	Genetic algorithms and other evolutionary methods would also be used as NAS search strategies, which generates a population of models and evolves them with mutation and other reproduction techniques.
	Evolutionary-based methods would produce promising results, although generally using too much GPU computation time (dozens to hundred GPU days), which would encourage research in other types of search strategies.
	
	Gradient-based (e.g. \cite{saxena2016convolutional, liu2018darts, pham2018efficient, chen2019progressive, weng2019automatic, xu2019pc}) NAS approaches aim a more continuous optimization, unlike the discrete and conditional from other types.
	A common idea is to present many possible operations (e.g. convolutions, poolings, skip-connections, etc) to be applied between one edge and another from a DAG representation and optimize weights representing what operation is better in that connection.
	This would represent not only a more flexible and fluid optimization, since it is more visible on how each operation contributes to the candidate score, but also works based on this approach that presents efficient results in less time.
	
	As for parameter estimation strategies, it mainly concerns evaluating validation error and training loss.
	Also, strategies to reduce candidate training time such as fewer epochs, reduced size, weight sharing and cosine annealing learning rate scheduler are employed. 
	Thus, evaluation of candidates can be realized in less computing time \cite{elsken2018neural}. 
	
	\subsection{Evolutionary-based NAS}
	
	Evolutionary-based NAS approaches generally consume a considerable amount of time, being infeasible to be employed in small research labs and companies, since it would cost a massive amount of GPU processing.
	Approaches which consume little GPU resources would be plausible to be employed in small companies and research labs, not only to popularize NAS application, but also to provide better models to specific-group problems.
	Although other techniques may be applied, some would be limited to a fixed search space.
	Not only that, but the already proposed evolutionary-based approaches may have focused too much in search with unlimited resource time, that other alternatives to employ a low cost evolutionary-based NAS may not be extensively studied.
	
	In virtue of these points, we propose an evolutionary-based NAS approach which focuses on obtaining efficient models with limited GPU time (24 hours in a NVIDIA GTX Titan Xp, a common card used in NAS benchmark) in a dynamic search space represented by GEP phenotype (tree-like structure).
	A NAS approach using only one GPU day may be used in groups with less resources.
	While this approach can be used with different types of deep networks, we focus this time only in convolutional networks (also called convnets), due to the broad range of experiments to be carried out.
	
	\section{Methodology}
	
	In this section, we defined the methodology and properties of our approach.
	
	\subsection{Workflow}
	
	Before entering in details of why and how each piece of NASGEP is used, we present its entire workflow below (being depicted in Figure \ref{fig:nasgep}):
	
	\begin{figure*}[hbt]
		\centering
		\includegraphics[width=0.6\textwidth]{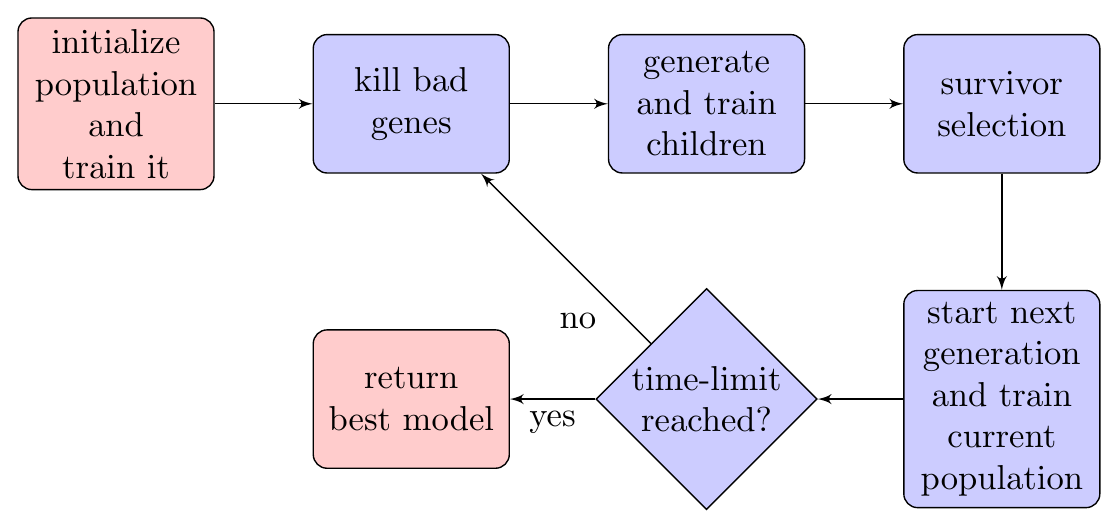}
		\caption{Overall representation of the NASGEP workflow.}
		\label{fig:nasgep}
	\end{figure*}
	
	\begin{enumerate}
		\item Genes for normal and reductions cells are generated (separately) for their initial population (size of $P_g$);
		\item An initial population of $P_r$ reduction cells is created;
		\item The first generation of $P_i$ models is generated and trained:
		\begin{enumerate}
			\item A random genotype (head and tail) is generated, to transform into the normal cell's phenotype;
			\item A random reduction cell is picked from the initial $P_r$ individuals;
			\item Then, each model is trained (only one epoch) to generated their fitness;
		\end{enumerate}
		\item Before reproduction of individuals, genes are killed and remaining are reproduced:
		\begin{enumerate}
			\item Genes not being used with a fitness below $T_g$ are discarded (to remove bad genes from population);
			\item Tournament selection is used to select two parents;
			\item At least $C^g_{min}$ children are created per generation;
			\item Children are generated until their size reaches $C^g_{max}$ in the generation or if the gene pool size reaches $P(max)_g$;
		\end{enumerate}
		\item The children generation is generated with tournament selection:
		\begin{enumerate}
			\item First, the new generation of reduction cells is created with reproduction of two parents (chosen by tournament selection, for each child);
			\item Then, for each individual in the normal population, reproduction is employed to generate their normal cell's genotype;
			\item For the individual's reduction cell generation, a tournament selection is applied to the population of reduction cells to select one of the alive cells;
		\end{enumerate} 		
		\item Children are trained:
		\begin{enumerate}
			\item Each child is trained for one epoch;
			\item If the fitness of a child reached a threshold $T_c$, the model is trained for another epoch (this is employed to reward better individuals in the search process);
			\item If a child reached the maximum number of epochs configured $E_{max}$, training is finished for this child and it is marked to be killed;
			\item In our specific approach, if a individual is marked to be killed after reaching the maximum number of epochs $E_{max}$, its current reduction cell is also marked to be killed (or not used for reproduction);
		\end{enumerate}
		\item After training of children population, NASGEP needs to reduce population size (from individuals and also reduction cells) -- thus applying survivor selection:
		\begin{enumerate}
			\item First, the best individual is always preserved -- elitism;
			\item Then, the oldest individual is killed;
			\item If any, dead individuals (which have already trained for $E_{max}$ epochs) are discarded;
			\item Finally, individuals in excess for both reduction and normal populations are removed based on their fitness ($P_i$ individuals including the best one are preserved);
			\item Reduction cells that are being used will survive to the next generation;	
			\item Thus, after survivor selection, more than $P_i$ individuals can be found in the reduction population;
			\item Then, generation is finished;
		\end{enumerate}
		\item For the next generation, each model is trained like step 6;
		\item Workflow goes back to step 4 and moves on, until the time-limit of 1 GPU day.
		
	\end{enumerate}
	
	\subsection{Why Gene Expression Programming?}
	
	The phenotype representation of genetic programming (and its variations, like gene expression programming \cite{ferreira2001gene}) consists of a binary tree with many leaves, which can represent the inputs of a convolutional cell and the root as its output (with intermediary nodes being convolutions and binary operations to join their output, like concatenation and addition).
	Thus, this phenotype provides a representation with little modifications to adapt for convolutional cells.
	
\begin{figure}[htb]
		\centering
		\hfill
		\begin{subfigure}[b]{0.17\textwidth}
			\centering
			\includegraphics[width=\textwidth]{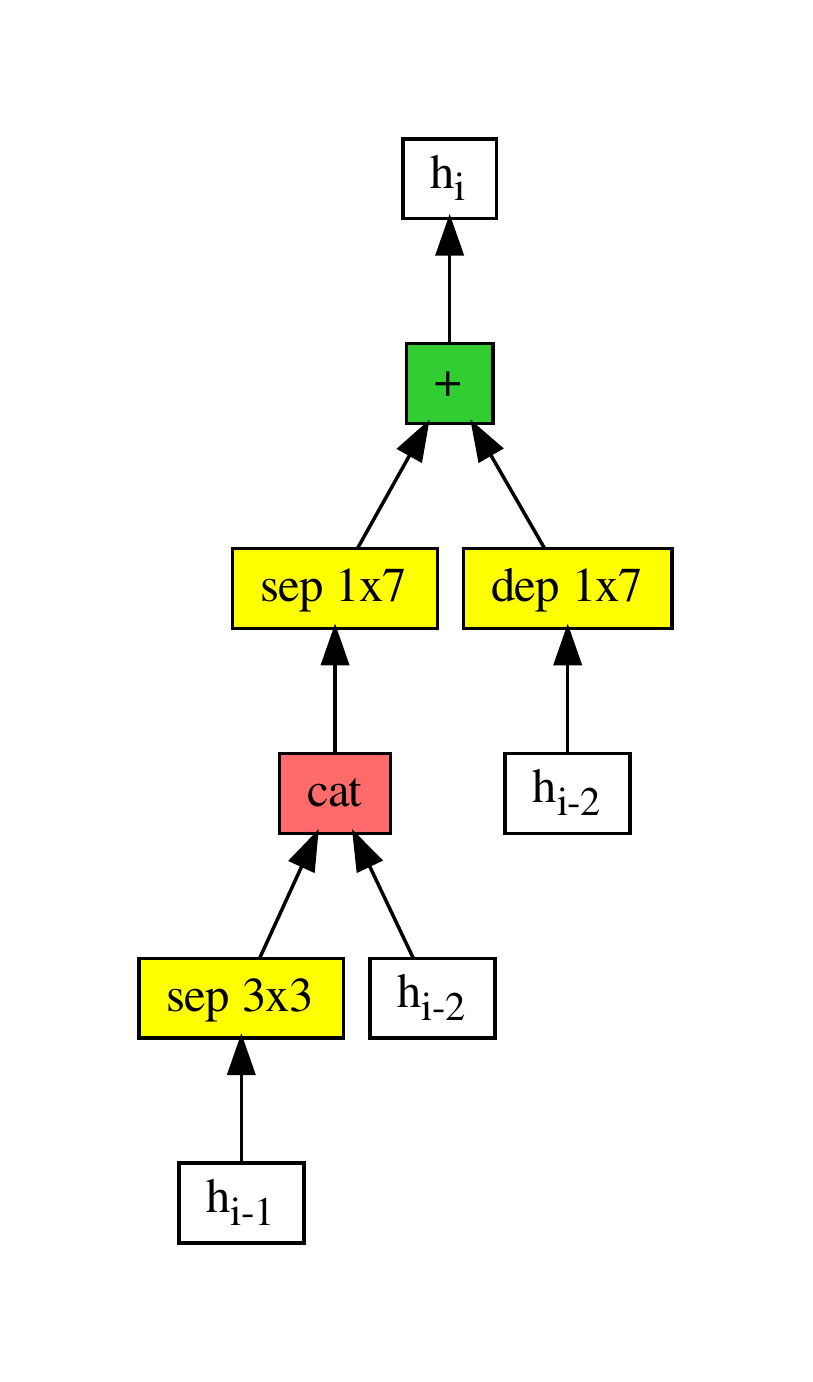}			
			\caption{Deep and shallow}
			\label{fig:example1}
		\end{subfigure}
		\hfill
		\begin{subfigure}[b]{0.3\textwidth}
			\centering
			\includegraphics[width=\textwidth]{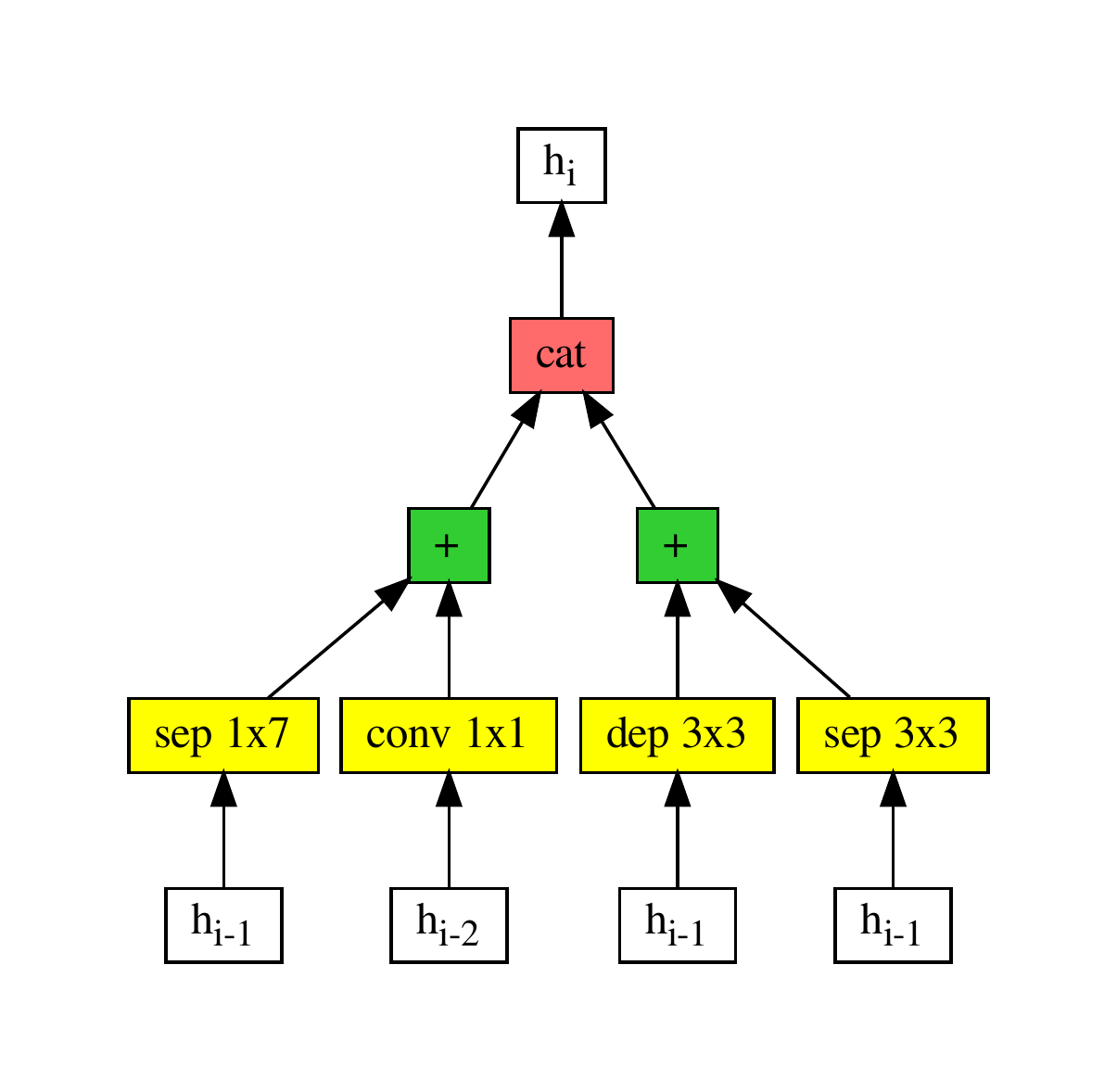}
			\caption{NAS cell-like representation}
			\label{fig:example2}
		\end{subfigure}
		\hfill
		\caption{Some examples of how a GEP phenotype can represent a convolutional cell. Leaves of this tree, i.e. outputs of previous cells ($h_{0}, ..., h_{i-2}, h_{i-1}$), are used as inputs of many operations (like convolutions, additions and concatenations). They are combined to produce the current cell's output $h_i$. This entire representation is repeated many times to generate the full convolutional network.}
		\label{fig:examples}
	\end{figure}

	Not only that, but this representation provides flexibility in searching different models.    
	For example: one branch is more deeper and the other more shallower (Figure \ref{fig:example1}); other with additions and concatenations alternating with each other, even simulating a NAS-cell based representation (Figure \ref{fig:example2}).
	
	Another point to be highlighted is the simple genotype representation in GEP. 
	Similar to genetic algorithm, genotype in GEP is a sequence of elements (e.g. string of characters and numbers).
	Not only modifications in this sequence can be easily done (like mutation and other reproduction methods), but the search space can be transferred to another search strategy.
	This simple sequence of head and tail may be changed with other techniques, like a Long Short-Term Memory (LSTM) architecture.
	Although the search strategy is modified, the dynamic search space representation remains intact.
	
	Also, using Automatically Defined Functions (ADFs, i.e. genes) encourages the optimization of small modules and combinations between them.
	Rather than optimize a bigger cell, optimizations on smaller modules would facilitate fitness convergence.
	If one part of a bigger cell interferes negatively on its fitness, the entire cell structure is discarded.
	But, if many of these small modules are found in different cells, through one cell may have lower fitness because of a bad gene, other good genes will obtain good fitness of other cells.
	Then, through generations, bad genes will be discarded and good ones will be combined in newer individuals.
	
	\subsection{Network Blocks: Functions and Terminals}
	
	In GEP, an individual is composed of both head and tail sequences.
	Both are combined to generate the individual's phenotype.
	For the head, functions and terminals (elements of a GEP sequence) are used.
	As for the tail, only terminals can be employed.
	The question is: what are functions and terminals in our NAS approach?
	
	For normal and reduction cells, the following elements are used as functions: addition and concatenation.
	For terminals, only ADFs (i.e. genes) are used.
	As for genes, addition and many types of convolutions (with different kernel sizes) are used as functions.
	For terminals, convolutions and previous cells' outputs. For the latter, only normal cells have these outputs as terminals. This was chosen since, in the reduction cell, additions with them would not be possible as they would have different feature map sizes.
	These elements are used to compose the cells in our network.
	Convolution blocks in this case are defined as the sequence of ReLU activation -$>$ Convolution -$>$ Batch Normalization, according \cite{he2016identity} (also followed in \cite{zoph2017learning}). 
	
	As for the types of convolutions, we used the following:
	\begin{itemize}
		\item Point-wise convolution (1x1 kernel);
		\item Depth-wise convolution with kernels 3x3, 5x5, 3x5, 5x3, 1x7 or 7x1;
		\item Separable depth-wise convolution (a point-wise convolution followed by a depth-wise) or in the inverse other, with kernel sizes of 3x3, 5x5, 3x5, 5x3, 1x7 and 7x1.
	\end{itemize}
	
	\subsection{Convolutional Network Peculiarities}
	
	With a diversified environment of convolutions, we can find useful combinations to represent a specific dataset.
	Also, as we are working with concatenations, the number of channels may increase on different occasions.
	Thus, a point-wise convolution is applied when a input encounter with another feature map with lower channel size (e.g. these two are inputs of an addition), or when the output of a cell is greater than the current size (only outputs of reduction cells change the number of channels, but also only double it).
	To reduce training time, whenever a convolution block or gene has validation accuracy greater than before, its current weights are saved to the disk.
	Then, whenever a new individual is created, these weights are loaded to the model, rather than being randomly initialized with, in our case, Kaiming initialization \cite{he2015delving}.
	
	\subsection{Reproduction}
	
	Generation of new individuals is executed in the reproduction step.
	The default techniques of GEP are used to generate them, being applied to a pair of parents the following (in order): mutation, transposition, root transposition, one-point and then two-point recombination.
	Default parameters from the original paper were employed.
	Also, this reproduction is also applied to genes, to generate and improve these modules for next generations.   
	One different thing in our approach is that, to encourage combination of addition and concatenation in different ways, elements in an individual's head only change to a function.
	
	\section{Experiments}
	
	Experiments were conducted in a NVIDIA GTX Titan Xp with 12 GB VRAM, a common card for GPU time evaluation.
	To assess the capability of our proposed work, experiments on common benchmark datasets were executed. 
	For their popularity in similar experiments, the CIFAR-10, CIFAR-100 and ImageNet datasets were used in this step.
	Each dataset is composed of common objects and living beings found in images.
	Also, different numbers of classes, samples per class and resolutions are employed in them.
	Before entering in more details regarding the experiments executed, hyper-parameters and other details about our architecture search are discussed in the next subsection. 
	
	\subsection{Search and full training phases' details}
	
	For the search phase, some parameters need to be defined differently from the full model training phase, since we want to evaluate as many models as possible.
	Thus, reduction of model size is applied for network candidates.
	The initial feature map size for the first convolution (before normal and reduction cells) is set to 16.
	Candidates are evaluated by their fitness.
	In our experiments the validation accuracy was employed as the fitness score.
	
	\begin{figure*}[hbt]
		\centering
		\includegraphics[width=0.99\textwidth]{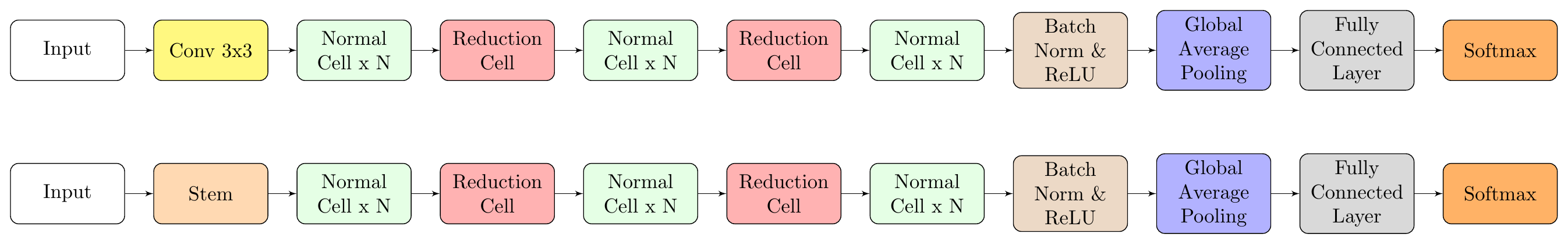}
		\caption{Overall network representation of CIFAR and ImageNet networks, respectively.}
		\label{fig:cifar_and_imagenet}
	\end{figure*}
	
	Initial population size $P_i$ for each generation is fixed to 10.
	For normal and reduction cells, the genotype's head size is equal to 4 (thus tail size to 5). 
	For their genes, head size was set to 1 (and tail size to 2).
	These values were chosen to enable a more flexible combination (e.g. vast combination of different small modules).
	The initial size of the gene pool is set to $P_g=50$ and its maximum size $P(max)_g$ is set to 100. For each generation, at least $C^g_{min} = 2$ genes are created and at most $C^g_{max} = 10$.
	$T_g$ is updated at the end of each generation to the $P_i$th best individual's fitness (for removal of bad genes).
	The threshold $T_c$ is updated to 75\% of this fitness.
	Number of parameters of network candidates was limited to 300 thousand to avoid generating models bigger than memory available.
	
	Training for each candidate can reach $E_{max} = 10$ epochs, using Stochastic Gradient Descent (SGD) with cosine annealing learning rate scheduler for optimization \cite{loshchilov2016sgdr}.
	Also, batch size is set to $512$ to focus on rapidly training.
	The initial learning rate value is set to $0.1$ for the search phase, to obtain quickly the validation error of each candidate that is close to the error after full training.
	In the full training phase, batch size is reduced to $128$ and learning rate to $0.025$, which focus on over-fitting reduction, besides its slow convergence.
	
	Number of repetitions for the normal cell (see Figure \ref{fig:cifar_and_imagenet} for reference) in each stage is set to $N=3$.
	In the full training phase, the initial feature map size is set to 64.
	The ImageNet's stem has three sequential 3x3 convolution blocks (where the first do not have ReLU activation) with stride $=2$.
	The first convolution block increases the number of channels to half the initial feature map size (size of 32) and the second convolution block to the initial size chosen (in this case, 64).
	Also, the number of epochs goes from 10 to 300 (for CIFAR datasets).
	The number of epochs of 90 is used in the ImageNet full training phase (no search phase is employed here).
	After the last normal cell, batch normalization followed by a ReLU activation is applied. 
	Then, global average pooling is employed to reduce the feature map size of each channel to 1x1.
	Finally, a fully connected layer reduces the output size to the number of classes, to be inputted in the softmax. 
	
	For data augmentation, mean and standard deviation correction was applied, after application of random crop and horizontal flip, CutOut \cite{devries2017improved} (length $=16$) and AutoAugment (AA) \cite{cubuk2018autoaugment}. For regularization, we also employed drop-path \cite{larsson2016fractalnet} with rate $=0.1$.  
	Weight decay of $0.0005$ and momentum of $0.9$ were also chosen values for regularization.
	
	\subsection{Experiments with CIFAR datasets}
	
	A popular benchmark dataset for image classification is the CIFAR project \cite{krizhevsky2009learning}, divided into two datasets: CIFAR-10 and CIFAR-100 (where the number is the amount of classes in the dataset).
	For both datasets, an amount of 50000 images for training is found. Testing subsets contain 10000 images each. Classes for both datasets have the same number of samples (6000 each class for CIFAR-10 and 600 for CIFAR-100).
	
	Search and training is employed in the CIFAR datasets separately.
	Not only that, but to assess the reliability of NASGEP, a total of five runs are done for both datasets (higher values would be infeasible, since searching and training deep learning models consume an excessive amount of resources).
	In the search phase (and also tuning of our method), the testing subset was not touched. 
	Thus, we extract 5000 samples from the training subset to be the validation subset, where we would find which combination of normal and reduction cells achieved superior fitness.
	
	\subsubsection{Ablation Study}
	
	In this study, we aim to analyze specific components (data augmentation and regularization) used to train our model. 
	Besides, evaluation of the weight sharing approach and training with random sampling is also exposed. 
	The objective is not only to determine which strategies have superior fitness, but also if our search proposal is really obtaining better models in the search space.
	Table \ref{tab:ablation} shows the experiments employed in this study.
	
	The first four experiments aim to analyze data augmentation and regularization in the same searched models. 
	The last one focuses on the search analysis using the best data augmentation configuration. 
	All experiments used CutOut.
	In general, combinations of these strategies improve accuracy satisfactory.
	Drop-path together with AA was an exception that, although had lower error than using only drop-path, it was worse than using only AA.
	This was probably caused by the inclusion of noise in the application of drop-path, which would limit the learning of some data variations generated by the data augmentation approach.
	
	To evaluate the different between random sampling and the searched models, we use the relative improvement (RI) metric in \cite{Yang2020NAS}, $RI = 100 \times(Acc_m - Acc_r)/Acc_r$, with $Acc_m$ being the mean of the five runs of our proposal and $Acc_r$ the mean of five random samplings.
	We obtained a $RI = 0.3098$, similar to DARTS ($RI=0.32$). 
	Since the combination of CutOut and AutoAugment  achieved the best results, they will be used in remaining experiments.
	
	\begin{table}[htb]
		\centering
		\caption{Ablation study in the CIFAR-10 dataset with different data augmentation methods, regularization and search strategies. }
		\label{tab:ablation}
		\begin{tabular}{@{}lcc@{}}
			\toprule
			& \textbf{Test error (in \%)} & \textbf{Params}  \\ \midrule
			CutOut only                     & $3.54\pm0.15$               & $3.62$M$\pm0.18$ \\
			\textbf{CutOut + AA}             & $\textbf{2.82}\pm\textbf{0.13}$               & $3.62$M$\pm0.18$ \\
			CutOut + Drop-Path               & $3.24\pm0.10$               & $3.62$M$\pm0.18$ \\
			CutOut + Drop-Path + AA          & $2.99\pm0.07$               & $3.62$M$\pm0.18$ \\ \midrule
			CutOut + AA + No Weight Sharing & $2.92\pm0.19$               & $3.48$M$\pm0.34$ \\
			CutOut + AA + Random Sampling   & $3.12\pm0.28$               & $2.54$M$\pm0.55$ \\ \bottomrule
		\end{tabular}
	\end{table}
	
	\subsubsection{Evaluation on the Best Strategy}
	
	In Table \ref{tab:cifar_results}, state-of-the-art manually designed and NAS approaches are compared if our own in CIFAR-10 and CIFAR-100 datasets.
	
	\begin{table*}[htb]
		\centering
		\caption{Results from CIFAR datasets.}
		\label{tab:cifar_results}{%
			\begin{tabular}{@{}lccccccc@{}}
				\toprule
				\multirow{2}{*}{\textbf{NAS}} & \multirow{2}{*}{\textbf{Search Space}} & \multirow{2}{*}{\textbf{Search Strategy}} & \multirow{2}{*}{\textbf{GPU Days}} & \multicolumn{2}{c}{\textbf{CIFAR-10}} & \multicolumn{2}{c}{\textbf{CIFAR-100}} \\
				&  &  &  & \textbf{Test error (in \%)} & \textbf{Params} & \textbf{Test error (in \%)} & \textbf{Params} \\ \midrule
				WideResNet \cite{zagoruyko2016wide} & Chain & Manual & N/A & $3.08\pm0.16$ & $36.5$M & $18.41\pm0.27$ & $36.5$M \\
				ResNeXt \cite{xie2017aggregated} & Multi-branch & Manual & N/A & $3.58$ & $34.4$M & $17.31$ & $34.4$M \\
				DenseNet-BC \cite{huang2017densely} & Chain & Manual & N/A & $3.46$ & $25.6$M & $17.18$ & $25.6$M \\ \midrule                
				NAS \cite{zoph2016neural} & Chain & RNN and RL & 2000 & $4.47$ & $7.1$M & - & - \\
				NASNet \cite{zoph2017learning} & NAS-cell & RL & 2000 & $2.65$ & $3.3$M & - & - \\
				AmoebaNet \cite{real2018regularized} & NAS-cell & Evolution & 3150 & $2.55$ & $2.8$M & - & - \\
				EANN-Net \cite{chen2019auto} & Chain & Evolution &  - & $7.05\pm0.02$ & - & - & - \\
				CoDeepNEAT \cite{miikkulainen2017evolving} & Multi-branch & Evolution & $>1$ & $7.3$ & - & - & - \\
				Genetic CNN \cite{xie2017genetic} & Multi-branch & Evolution & 20 & $7.10$ & - & $29.03$ & - \\
				Hill-climbing \cite{elsken2017simple} & Multi-branch & Hill-climbing & 1 & $5.2$ & $19.7$M & $23.4$ & $22.3$M \\                               
				DARTS \cite{liu2018darts} & DAG-cell & Gradient & 4 & $2.76\pm0.09$ & $3.3$M & - & - \\ \midrule
				NASGEP & GEP-based cell & Evolution & 1 & $2.82\pm0.13$ & $3.62$M$\pm0.18$ & $18.83\pm0.39$ & $3.59$M$\pm0.30$ \\ \bottomrule
			\end{tabular}%
		}
	\end{table*}
	
\begin{figure}[htb]
		\centering
		\includegraphics[width=0.8\linewidth]{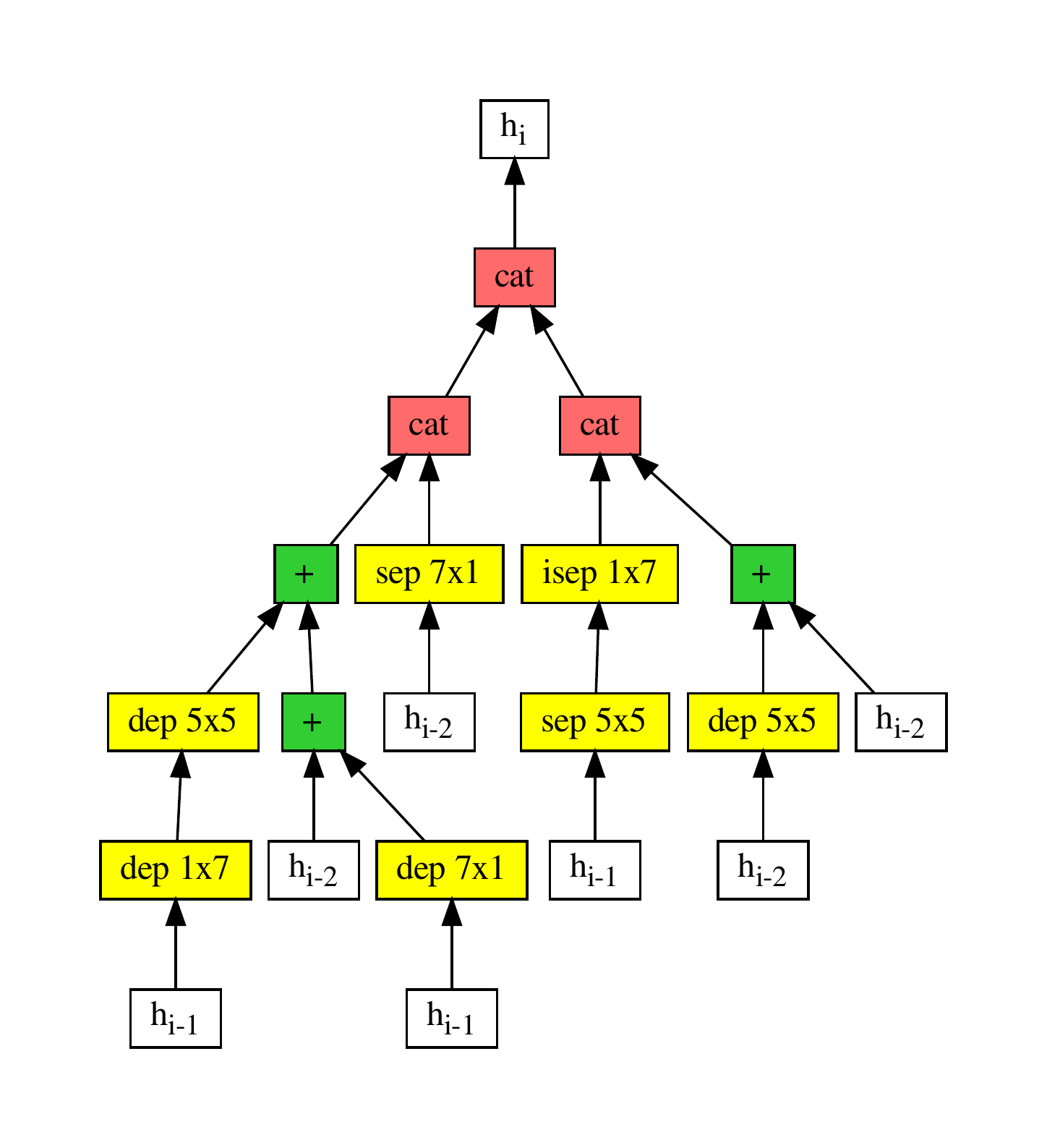}
		\caption{Normal cell from the best model found in CIFAR-10, also transferred to ImageNet mobile setting training.}
		\label{fig:cifar10_normal}
	\end{figure}

\begin{figure}[htb]
		\centering
		\includegraphics[width=\linewidth]{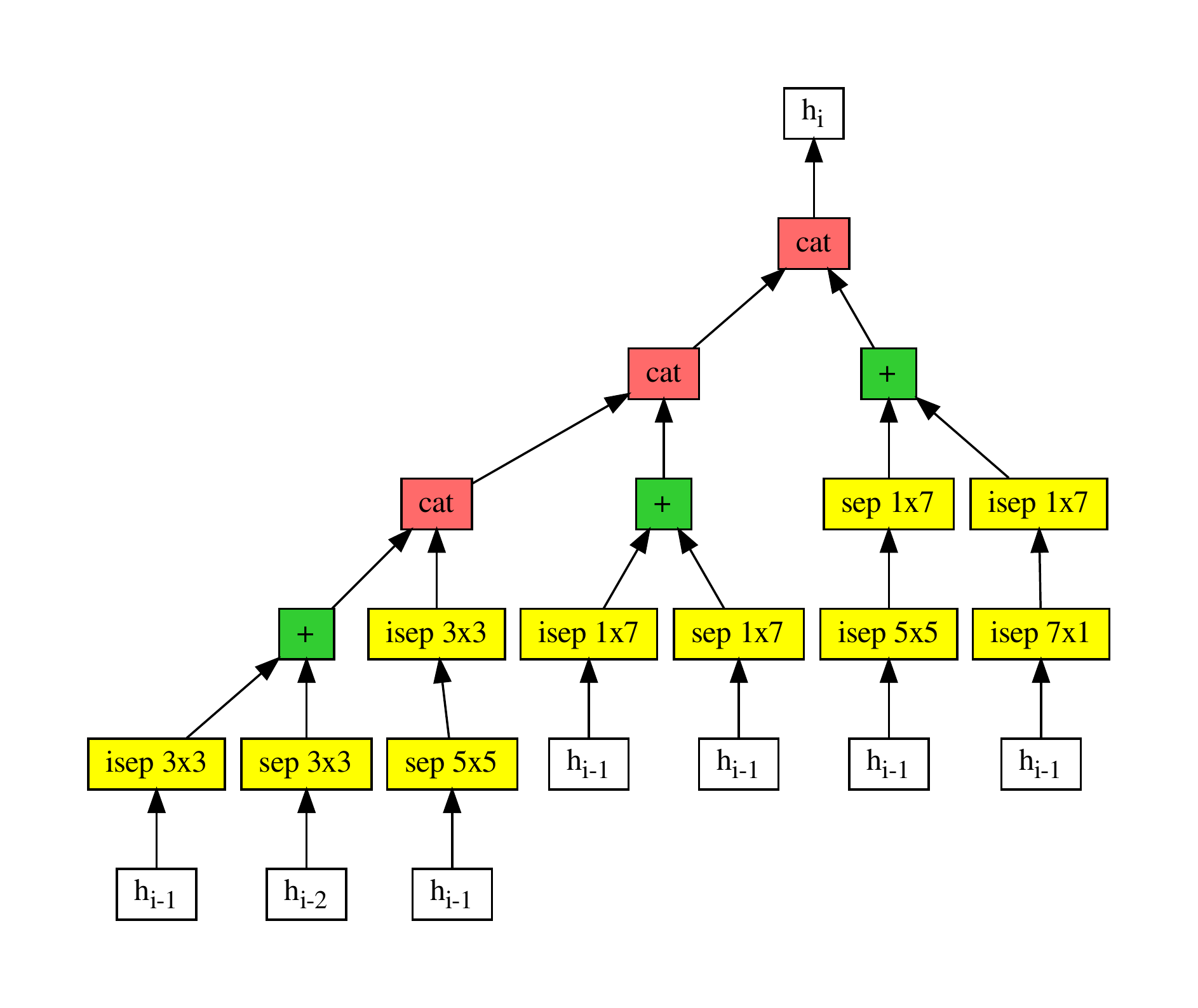}	
		\caption{Reduction cell from CIFAR-10's best model, which the convolutions following an input ($h_{i-1}$ or $h_{i-2}$) have stride $=2$ for feature map size reduction.}
		\label{fig:cifar10_reduction}        
	\end{figure}

	After five runs, we achieved a test error of $2.82\pm0.13$ in CIFAR-10, and $18.83\pm0.39$ in CIFAR-100, with our best models in each dataset obtaining an error of $2.67\%$ and $18.16\%$, respectively.
	Normal and reduction cells from the best candidate in CIFAR-10 can be seen in Figures \ref{fig:cifar10_normal} and \ref{fig:cifar10_reduction}, respectively.
	The structure of both cells are very different, also in the number of convolutional operations.
	Normal had a quantity similar to AmoebaNet cells.
	On the other hand, the reduction cell had a greater number of convolutions, mainly inverse separable convolutions. 
	Not only were they different between them, but also different between other approaches, which had more fixed structures, mainly in not combining addition and concatenation dynamically.
	Even with a flexible search space, our method obtained satisfactory cells for the final convnet.

	
	Number of parameters from candidates chosen were around three million, similar to other recent NAS approaches which focused on this reduction without losing their performance.
	Compared with the best NAS approaches like AmoebaNet and DARTS, NASGEP had only little worse results, showing the capability of the approach even with a diverse search space.
	Furthermore, NASGEP obtained significant results using only one GPU day (in contrast with the three thousand from AmoebaNet), and outperformed substantially other evolutionary-based approaches, seeing in Table \ref{tab:cifar_results}.  
	
	\subsection{Experiments with ImageNet on Mobile Setting}
	
	Another popular dataset used for image classification (and other problems such as localization and detection) is the ImageNet Large Scale Visual Recognition Challenge (ILSVRC) dataset \cite{deng2009imagenet}.
	This specific dataset contains more than one million training images from 1000 classes.
	Its validation and testing subsets contain 50000 images each.
	Since the ImageNet dataset contains a massive amount of images, only one run was executed for it.
	Similar to other NAS works (e.g. DARTS and AmoebaNet), we transferred the learned CIFAR cell for evaluation of the ImageNet case.
	Table \ref{tab:imagenet_results} shows results in both manually-designed models and NAS-generated ones. 
	Also, the result obtained with NASGEP is presented.    
	
	\begin{table*}[htb]
		\centering
		\caption{Results from ImageNet dataset in mobile setting.}
		\label{tab:imagenet_results}{%
			\begin{tabular}{@{}lcccccc@{}}
				\toprule
				\textbf{NAS} & \textbf{Search Space} & \textbf{Search Strategy} & \textbf{GPU Days} & \textbf{Top-1 error (in \%)} & \textbf{Top-5 error (in \%)} & \textbf{Params} \\ \midrule
				AlexNet \cite{krizhevsky2012imagenet} & Chain & Manual & N/A & $37.5$ & $17$ & $60$M \\
				SqueezeNet \cite{iandola2016squeezenet} & Chain & Manual & N/A & $39.6$ & $17.5$ & $0.42$M \\
				ShuffleNet \cite{zhang2018shufflenet} & Chain & Manual & N/A & $24.7$ & - & $5$M \\
				MobileNet \cite{howard2017mobilenets} & Chain & Manual & N/A & $29.4$ & - & $4.2$M \\ \midrule
				NASNet \cite{zoph2017learning} & NAS-cell & RL & 2000 & $26$ & $8.4$ & $5.3$M \\
				AmoebaNet \cite{real2018regularized} & NAS-cell & Evolution & 3150 & $24.3$ & $7.6$ & $6.4$M \\
				PNAS \cite{liu2017progressive} & NAS-cell & SMBO & 225 & $25.8$ & $8.1$ & $5.1$M  \\                
				DARTS \cite{liu2018darts} & DAG-cell & Gradient & 4 & $26.7$ & $8.7$ & $4.7$M \\ \midrule
				NASGEP & GEP-based cell & Evolution & 1 & $29.51$ & $10.37$ & $4.1$M \\ \bottomrule
			\end{tabular}%
		}
	\end{table*}
	
	In this case, NASGEP had a top-1 error of $29.51\%$ and top-5 of $10.37\%$, significantly greater than other approaches shown in Table \ref{tab:imagenet_results}. 
	Besides the fact that lower GPU resource time was used, it is crucial to further investigate alternatives to address this, even if the approaches compared already had slightly better results in CIFAR experiments (although we did achieve satisfactory results compared to AlexNet and SqueezeNet).
	
	\section{Conclusion}
	
	In this paper, an evolutionary-based neural architecture search, NASGEP, for limited GPU usage was developed.
	It mainly consisted of using GEP as its search space and strategy.
	Not only evaluation of limited GPU usage was in mind, but also to adapt the phenotype representation of GEP to a convolutional cell.
	Choosing this phenotype was mainly for the introduction of a more dynamic search space in NAS, but also to assess the reliability of employing evolutionary-based approaches in time-limited resources.
	Although our approach did not surpass state-of-the-art methods in CIFAR and ImageNet mobile settings, promising results were found, besting state-of-the-art manually-designed and NAS models in CIFAR-10 dataset (and similar to them in CIFAR-100 and ImageNet). 
	With this in regard, studies in evolutionary-based approaches with cell-based dynamic search spaces may be further pursued in time-limited GPU execution.
	
	Further research in how to take advantage of the GEP phenotype and variants may be pursued, aiming for more optimized search spaces.
	In this study, ADFs were only treated as terminals.
	Although this stabilizes the fitness for a specific gene (as limited the gene to be only added or concatenated in a cell), it reduces the ways a gene can be treated, such as being the input or output of another gene.
	Treating genes as functions may present a new environment for study and optimization: the way a gene is treated in a cell can somehow improve the performance of a model?
	Also, further experiments in different datasets but with the same classes would assess the capability of the best cells found to generalize to similar datasets.
	
	\section*{Acknowledgment}
	
	We would like to thank NVIDIA Corporation with the donation of the Titan Xp GPU used in our experiments.
	
	\bibliographystyle{IEEEtran}
	\bibliography{paper}

\begin{thebibliography}{10}
\providecommand{\url}[1]{#1}
\csname url@samestyle\endcsname
\providecommand{\newblock}{\relax}
\providecommand{\bibinfo}[2]{#2}
\providecommand{\BIBentrySTDinterwordspacing}{\spaceskip=0pt\relax}
\providecommand{\BIBentryALTinterwordstretchfactor}{4}
\providecommand{\BIBentryALTinterwordspacing}{\spaceskip=\fontdimen2\font plus
\BIBentryALTinterwordstretchfactor\fontdimen3\font minus
  \fontdimen4\font\relax}
\providecommand{\BIBforeignlanguage}[2]{{%
\expandafter\ifx\csname l@#1\endcsname\relax
\typeout{** WARNING: IEEEtran.bst: No hyphenation pattern has been}%
\typeout{** loaded for the language `#1'. Using the pattern for}%
\typeout{** the default language instead.}%
\else
\language=\csname l@#1\endcsname
\fi
#2}}
\providecommand{\BIBdecl}{\relax}
\BIBdecl

\bibitem{elsken2018neural}
T.~Elsken, J.~H. Metzen, and F.~Hutter, ``Neural architecture search: A
  survey,'' \emph{arXiv preprint arXiv:1808.05377}, 2018.

\bibitem{elsken2017simple}
T.~Elsken, J.-H. Metzen, and F.~Hutter, ``Simple and efficient architecture
  search for convolutional neural networks,'' \emph{arXiv preprint
  arXiv:1711.04528}, 2017.

\bibitem{elsken1804efficient}
T.~Elsken, J.~H. Metzen, and F.~Hutter, ``Efficient multi-objective neural
  architecture search via lamarckian evolution,'' \emph{ArXiv e-prints}, 1804.

\bibitem{szegedy2017inception}
C.~Szegedy, S.~Ioffe, V.~Vanhoucke, and A.~A. Alemi, ``Inception-v4,
  inception-resnet and the impact of residual connections on learning.'' in
  \emph{AAAI}, vol.~4, 2017, p.~12.

\bibitem{liu2018darts}
H.~Liu, K.~Simonyan, and Y.~Yang, ``Darts: Differentiable architecture
  search,'' \emph{arXiv preprint arXiv:1806.09055}, 2018.

\bibitem{zoph2017learning}
B.~Zoph, V.~Vasudevan, J.~Shlens, and Q.~V. Le, ``Learning transferable
  architectures for scalable image recognition,'' \emph{arXiv preprint
  arXiv:1707.07012}, vol.~2, no.~6, 2017.

\bibitem{real2018regularized}
E.~Real, A.~Aggarwal, Y.~Huang, and Q.~V. Le, ``Regularized evolution for image
  classifier architecture search,'' \emph{arXiv preprint arXiv:1802.01548},
  2018.

\bibitem{baker2016designing}
B.~Baker, O.~Gupta, N.~Naik, and R.~Raskar, ``Designing neural network
  architectures using reinforcement learning,'' \emph{arXiv preprint
  arXiv:1611.02167}, 2016.

\bibitem{liu2017hierarchical}
H.~Liu, K.~Simonyan, O.~Vinyals, C.~Fernando, and K.~Kavukcuoglu,
  ``Hierarchical representations for efficient architecture search,''
  \emph{arXiv preprint arXiv:1711.00436}, 2017.

\bibitem{real2017large}
E.~Real, S.~Moore, A.~Selle, S.~Saxena, Y.~L. Suematsu, J.~Tan, Q.~V. Le, and
  A.~Kurakin, ``Large-scale evolution of image classifiers,'' in
  \emph{International Conference on Machine Learning}, 2017, pp. 2902--2911.

\bibitem{miikkulainen2017evolving}
R.~Miikkulainen, J.~Liang, E.~Meyerson, A.~Rawal, D.~Fink, O.~Francon, B.~Raju,
  H.~Shahrzad, A.~Navruzyan, N.~Duffy \emph{et~al.}, ``Evolving deep neural
  networks,'' \emph{arXiv preprint arXiv:1703.00548}, 2017.

\bibitem{suganuma2017genetic}
M.~Suganuma, S.~Shirakawa, and T.~Nagao, ``A genetic programming approach to
  designing convolutional neural network architectures,'' in \emph{Proceedings
  of the Genetic and Evolutionary Computation Conference}.\hskip 1em plus 0.5em
  minus 0.4em\relax ACM, 2017, pp. 497--504.

\bibitem{zhong2018practical}
Z.~Zhong, J.~Yan, W.~Wu, J.~Shao, and C.-L. Liu, ``Practical block-wise neural
  network architecture generation,'' in \emph{Proceedings of the IEEE
  Conference on Computer Vision and Pattern Recognition}, 2018, pp. 2423--2432.

\bibitem{zoph2016neural}
B.~Zoph and Q.~V. Le, ``Neural architecture search with reinforcement
  learning,'' \emph{arXiv preprint arXiv:1611.01578}, 2016.

\bibitem{saxena2016convolutional}
S.~Saxena and J.~Verbeek, ``Convolutional neural fabrics,'' in \emph{Advances
  in Neural Information Processing Systems}, 2016, pp. 4053--4061.

\bibitem{pham2018efficient}
H.~Pham, M.~Y. Guan, B.~Zoph, Q.~V. Le, and J.~Dean, ``Efficient neural
  architecture search via parameter sharing,'' in \emph{ICML}, 2018.

\bibitem{chen2019progressive}
X.~Chen, L.~Xie, J.~Wu, and Q.~Tian, ``Progressive differentiable architecture
  search: Bridging the depth gap between search and evaluation,'' in
  \emph{Proceedings of the IEEE International Conference on Computer Vision},
  2019, pp. 1294--1303.

\bibitem{weng2019automatic}
Y.~Weng, T.~Zhou, L.~Liu, and C.~Xia, ``Automatic convolutional neural
  architecture search for image classification under different scenes,''
  \emph{IEEE Access}, vol.~7, pp. 38\,495--38\,506, 2019.

\bibitem{xu2019pc}
Y.~Xu, L.~Xie, X.~Zhang, X.~Chen, G.-J. Qi, Q.~Tian, and H.~Xiong, ``Pc-darts:
  Partial channel connections for memory-efficient differentiable architecture
  search,'' \emph{arXiv preprint arXiv:1907.05737}, 2019.

\bibitem{ferreira2001gene}
C.~Ferreira, ``Gene expression programming: A new adaptive algorithm for
  solving problems,'' \emph{Complex Systems}, vol.~13, no.~2, pp. 87--129,
  2001.

\bibitem{he2016identity}
K.~He, X.~Zhang, S.~Ren, and J.~Sun, ``Identity mappings in deep residual
  networks,'' in \emph{European conference on computer vision}.\hskip 1em plus
  0.5em minus 0.4em\relax Springer, 2016, pp. 630--645.

\bibitem{he2015delving}
------, ``Delving deep into rectifiers: Surpassing human-level performance on
  imagenet classification,'' in \emph{Proceedings of the IEEE international
  conference on computer vision}, 2015, pp. 1026--1034.

\bibitem{loshchilov2016sgdr}
I.~Loshchilov and F.~Hutter, ``Sgdr: Stochastic gradient descent with warm
  restarts,'' \emph{arXiv preprint arXiv:1608.03983}, 2016.

\bibitem{devries2017improved}
T.~DeVries and G.~W. Taylor, ``Improved regularization of convolutional neural
  networks with cutout,'' \emph{arXiv preprint arXiv:1708.04552}, 2017.

\bibitem{cubuk2018autoaugment}
E.~D. Cubuk, B.~Zoph, D.~Mane, V.~Vasudevan, and Q.~V. Le, ``Autoaugment:
  Learning augmentation policies from data,'' \emph{arXiv preprint
  arXiv:1805.09501}, 2018.

\bibitem{larsson2016fractalnet}
G.~Larsson, M.~Maire, and G.~Shakhnarovich, ``Fractalnet: Ultra-deep neural
  networks without residuals,'' \emph{arXiv preprint arXiv:1605.07648}, 2016.

\bibitem{krizhevsky2009learning}
A.~Krizhevsky and G.~Hinton, ``Learning multiple layers of features from tiny
  images,'' Citeseer, Tech. Rep., 2009.

\bibitem{Yang2020NAS}
\BIBentryALTinterwordspacing
A.~Yang, P.~M. Esperança, and F.~M. Carlucci, ``Nas evaluation is
  frustratingly hard,'' in \emph{International Conference on Learning
  Representations}, 2020. [Online]. Available:
  \url{https://openreview.net/forum?id=HygrdpVKvr}
\BIBentrySTDinterwordspacing

\bibitem{zagoruyko2016wide}
S.~Zagoruyko and N.~Komodakis, ``Wide residual networks,'' \emph{arXiv preprint
  arXiv:1605.07146}, 2016.

\bibitem{xie2017aggregated}
S.~Xie, R.~Girshick, P.~Doll{\'a}r, Z.~Tu, and K.~He, ``Aggregated residual
  transformations for deep neural networks,'' in \emph{Computer Vision and
  Pattern Recognition (CVPR), 2017 IEEE Conference on}.\hskip 1em plus 0.5em
  minus 0.4em\relax IEEE, 2017, pp. 5987--5995.

\bibitem{huang2017densely}
G.~Huang, Z.~Liu, K.~Q. Weinberger, and L.~van~der Maaten, ``Densely connected
  convolutional networks,'' in \emph{Proceedings of the IEEE conference on
  computer vision and pattern recognition}, vol.~1, no.~2, 2017, p.~3.

\bibitem{chen2019auto}
Z.~Chen, Y.~Zhou, and Z.~Huang, ``Auto-creation of effective neural network
  architecture by evolutionary algorithm and resnet for image classification,''
  in \emph{2019 IEEE International Conference on Systems, Man and Cybernetics
  (SMC)}.\hskip 1em plus 0.5em minus 0.4em\relax IEEE, 2019, pp. 3895--3900.

\bibitem{xie2017genetic}
L.~Xie and A.~Yuille, ``Genetic cnn,'' in \emph{Proceedings of the IEEE
  international conference on computer vision}, 2017, pp. 1379--1388.

\bibitem{deng2009imagenet}
J.~Deng, W.~Dong, R.~Socher, L.-J. Li, K.~Li, and L.~Fei-Fei, ``Imagenet: A
  large-scale hierarchical image database,'' in \emph{2009 IEEE conference on
  computer vision and pattern recognition}.\hskip 1em plus 0.5em minus
  0.4em\relax Ieee, 2009, pp. 248--255.

\bibitem{krizhevsky2012imagenet}
A.~Krizhevsky, I.~Sutskever, and G.~E. Hinton, ``Imagenet classification with
  deep convolutional neural networks,'' in \emph{Advances in neural information
  processing systems}, 2012, pp. 1097--1105.

\bibitem{iandola2016squeezenet}
F.~N. Iandola, S.~Han, M.~W. Moskewicz, K.~Ashraf, W.~J. Dally, and K.~Keutzer,
  ``Squeezenet: Alexnet-level accuracy with 50x fewer parameters and< 0.5 mb
  model size,'' \emph{arXiv preprint arXiv:1602.07360}, 2016.

\bibitem{zhang2018shufflenet}
X.~Zhang, X.~Zhou, M.~Lin, and J.~Sun, ``Shufflenet: An extremely efficient
  convolutional neural network for mobile devices,'' in \emph{Proceedings of
  the IEEE conference on computer vision and pattern recognition}, 2018, pp.
  6848--6856.

\bibitem{howard2017mobilenets}
A.~G. Howard, M.~Zhu, B.~Chen, D.~Kalenichenko, W.~Wang, T.~Weyand,
  M.~Andreetto, and H.~Adam, ``Mobilenets: Efficient convolutional neural
  networks for mobile vision applications,'' \emph{arXiv preprint
  arXiv:1704.04861}, 2017.

\bibitem{liu2017progressive}
C.~Liu, B.~Zoph, J.~Shlens, W.~Hua, L.-J. Li, L.~Fei-Fei, A.~Yuille, J.~Huang,
  and K.~Murphy, ``Progressive neural architecture search,'' \emph{arXiv
  preprint arXiv:1712.00559}, 2017.

\end{thebibliography}
	
\end{document}